\documentclass{article} % For LaTeX2e
\usepackage{nips13submit_e,times}
\usepackage{hyperref}
\usepackage{url}
\usepackage{graphicx}
\usepackage{bbm}
\usepackage{amsfonts}
\usepackage{amsthm}
\usepackage{amsmath}

\title{Supervised Learning with \\ Indefinite Topological Kernels}

\author{
Tullia Padellini\\
Dipartimento di Scienze Statistiche\\
Sapienza - Universit\`a di Roma\\
\texttt{tullia.padellini@uniroma1.it} \\
\And
Pierpaolo Brutti \\
Dipartimento di Scienze Statistiche\\
Sapienza - Universit\`a di Roma\\
\texttt{pierpaolo.brutti@uniroma1.it} \\
}

\newcommand{\norm}[1]{\left\lVert#1\right\rVert}
\newtheorem{proposition}{Proposition}[section]
\newtheorem{theorem}{Theorem}[section]

\newtheorem{lemma}[theorem]{Lemma}

\theoremstyle{definition}
\newtheorem{definition}{Definition}[section]

\nipsfinalcopy % Uncomment for camera-ready version

\begin{document}

\maketitle

\begin{abstract}

Topological Data Analysis (\texttt{TDA}) is a recent and growing branch of statistics devoted to the study of the shape of the data. In this work we investigate the predictive power of \texttt{TDA} in the context of supervised learning. Since topological summaries, most noticeably the Persistence Diagram, are typically defined in complex spaces, we adopt a kernel approach to translate them into more familiar vector spaces. We define a topological exponential kernel, we characterize it, and we show that, despite not being positive semi--definite, it can be successfully used in regression and classification tasks.
\end{abstract}

\section{Topological Data Analysis / Motivation}

As we are dealing with increasingly complex data, our need for characterizing them through a few, interpretable features has grown considerably. Topology has proven to be a useful tool in this quest for ``insights on the data'', since it characterizes objects through their connectivity structure, i.e. connected components, loops and voids. In a statistical framework, this characterization yields relevant information: connected components correspond to clusters \cite{Chazal2013} while loops represent periodic structures \cite{Perea2015}.
At the crossroad between Computational Topology and Statistics, Topological Data Analysis  $\texttt{TDA}$ consists of techniques aimed at recovering the topological structure of data. Although topology has always been considered a very abstract branch of mathematics, it has some properties that are extremely desirable in data analysis, such as: 
\begin{itemize}
\item \emph{It does not depend on the coordinates of the data, but only on pairwise distances.} In many applications, coordinates are not given to us or, even if they are, they have no meaning and they could be misleading. 
\item \emph{It is invariant with respect to a large class of deformations.} Two object that can be deformed into one another without cutting or gluing are topologically  equivalent, meaning that topological methods are flexible.
\item \emph{It allows for a discrete representation of the objects we study.} Most continuous objects can be approximated with a discrete but topologically equivalent object, for which it is easier to define algorithms.
\end{itemize}

Even though the impressive growth of \texttt{TDA} literature in the last couple of years has yield several inference--ready tools, this hype has not yet been matched by popularity in the practice of data analysis, and in the applied works \texttt{TDA} is still mostly used as an exploratory step only. Our goal is to show that topological summaries can be useful in inferential tasks as well, with a special focus on supervised learning. In order to do so, we introduce a new family of Topological kernels, we investigate its properties and finally we show some real data application to classification and regression problems.

\subsection{Persistent Homology Groups}

The goal of \texttt{TDA} is to recover the topological structure of some object  $\mathbb{X} $ by estimating topological invariants such as Homology Groups, Euler Characteristics, Betti numbers and other standard concepts of Algebraic Topology (we refer to \cite{Hatcher2002} for a complete survey).

If the object we are interested in are data coming in the form of point--cloud  $\mathbb{X}_n = \{ X_1,\ldots,X_n \} $, however, it is not possible to compute these invariants directly, or, even if it is, they retain no relevant information. A point--cloud  $\mathbb{X}_n $, in fact, has a trivial topological structure per se, as it is composed of as many connected components as there are observations and no higher dimensional features. The first step in the \texttt{TDA} pipeline thus consists in enriching the data by encoding them into a filtration, for example by
growing each point  $X_i $ into a ball 
\[
B(X_i, \varepsilon) = \big\{ x \,\big|\, d_{\mathbb{X}} (x, X_i) \leq \varepsilon \big\},
\]
of fixed radius  $\varepsilon$. As long as  $(\mathbb{X}_n, d_{\mathbb{X}})$ is a metric space, any arbitrary distance  $d_{\mathbb{X}}$ can be used to define  $B(X_i, \varepsilon) $. The metric  $d_{\mathbb{X}}$ can be used to enforce some desired property, for example define a distance function, the \emph{Distance to Measure} to \emph{robustify} the estimate \cite{chazal2014robust}.

At each resolution  $\varepsilon$, the \emph{cover}
\[
\mathbb{X}_n^{\varepsilon} = \bigcup_{i=1}^n B(X_i, \varepsilon),
\]
has a different topological structure. When  $\varepsilon$ is very small,  $\mathbb{X}^{\varepsilon}_n$ is topologically equivalent to $\mathbb{X}_n$; as  $\varepsilon$ grows, however, balls of the cover start to intersect, ``giving birth'' to loops, voids and other topologically interesting structures. At some point, when connected components merge, loops are filled and so on, these structures start to ``die''. Eventually when  $\varepsilon$ reaches the diameter of the data  $\mathbb{X}_n$,  $\mathbb{X}_n^{\varepsilon}$ is \emph{contractible}, or in other words, topologically equivalent to a filled ball, and again retains no information. 

\begin{figure}
\centering
\includegraphics[width = 1\textwidth]{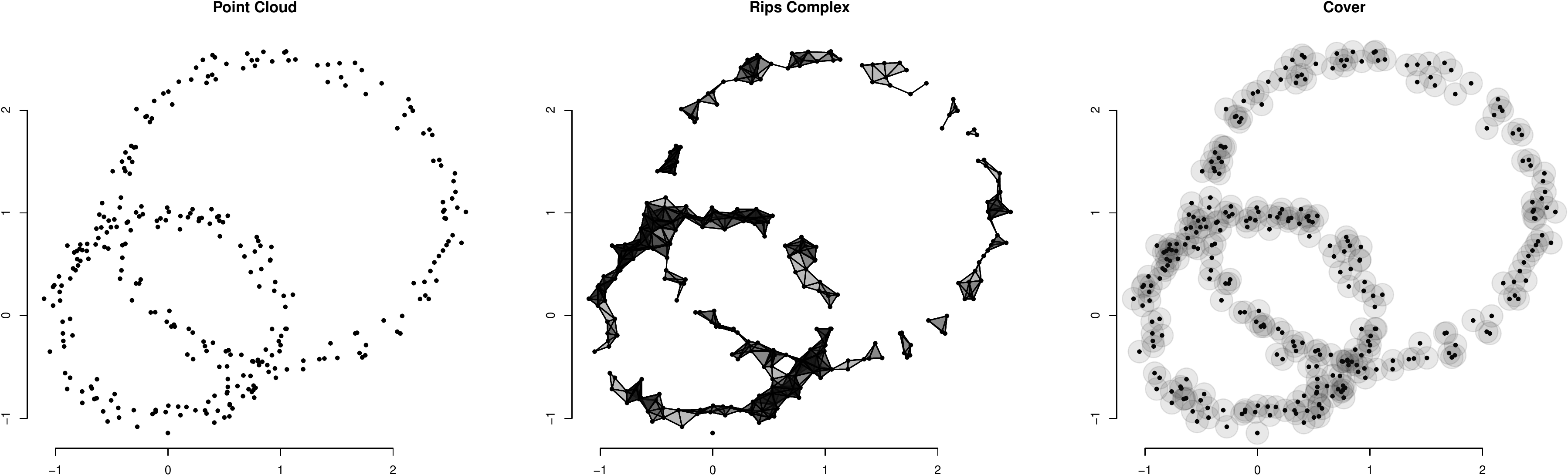}
\caption{From left to right: Data $\mathbb{X}$, Rips complex $Rips_{\varepsilon}(\mathbb{X})$ and corresponding cover $\mathbb{X}^{\varepsilon}$.}
\label{fig:toydata}
\end{figure}

The main feature of the cover is that different levels,  $\mathbb{X}_n^{\delta}$, and $\mathbb{X}_n^{\eta}$ say, are related by inclusion:  $\mathbb{X}_n^{\delta} \subseteq \mathbb{X}_n^{\eta}$ when  $\delta \leq \eta$. Formally  $\mathcal{F}=\{\mathbb{X}_n^{\varepsilon}\}_{\varepsilon}$ is the sublevel set \emph{filtration} of the distance function  $d_{\mathbb{X}}$ and the map $i^{\eta}_{\delta}: \mathbb{X}^{\delta} \mapsto \mathbb{X}^{\eta}$ is called \emph{inclusion map}. The key point of encoding data into a filtration is that we can track the evolution of each feature and see when it appears and disappears. Until now we have described a very general framework, in which we denoted as ``feature'' an arbitrary topological invariant; from here on we will always implicitly refer to Homology Groups  $H_k (\mathbb{X})$, following most of the \texttt{TDA} literature (with noticeable exceptions in \cite{Bobrowski2017, Giusti2015}).

Roughly speaking, at every level $\varepsilon$ of the filtration $\mathcal{F}$ each element of the dimension $0$ Homology group of $\mathbb{X}^{\varepsilon}$,  $H_0(\mathbb{X^{\varepsilon}})$,
represents a connected component of  $\mathbb{X}$, each element of the dimension  $1 $ Homology group, $H_1(\mathbb{X}^{\varepsilon})$, represents a loop, the dimension $2$ Homology group contains voids and so on.

In order to understand which  $k$--dimensional feature survives between $\eta$ and  $\delta$, it is necessary to build the map
\[
f: H_k(\mathbb{X}^{\eta}) \mapsto H_k(\mathbb{X}^{\delta}),
\] 
that shows how homology groups at  $\eta$ and  $\delta$ are related. However, since Homology is a \emph{functor}, it is easy to see that it induces a linear map $H(i_{\eta} ^{ \delta}) : H(\mathbb{X}^{\eta})\mapsto H(\mathbb{X}^{\delta})$ on the inclusion map of the $\mathbb{X}^{\eta}\hookrightarrow\mathbb{X}^{\delta} $, so that $f = H(i_t ^s) $. This is the basic idea behind the notion of
Persistent Homology Groups, a \emph{multiscale} version of Homology Groups that analyses the evolution of the topology of the elements of a filtration.\\

\begin{definition}[Persistent Homology Groups]
 Given a filtration $\mathcal{F} = \{\mathbb{X}^{\varepsilon}\}_\varepsilon$ indexed on $\mathbb{R}$, i.e.~a sequence of topological spaces $\mathbb{X}^{\varepsilon}$ for each $\varepsilon \in \mathbb{R}$ and maps  $\mathbb{X}^{\eta}\hookrightarrow \mathbb{X}^{\delta}$ for $\eta\leq \delta$, there are natural maps
\[
H(i_\eta ^{\delta}) : H_k (\mathbb{X}^{\eta}) \mapsto H_k(\mathbb{X}^{\delta}),
\]
induced by functoriality. We define the dimension--$k$ Persistent
Homology Group  $H_{k,p} $, where  $p = \delta-\eta $, as the image of the induced map  $H(i_\eta^{\delta})$.
\end{definition}

Persistent Homology Groups $H_{k,p}$ can be defined for every dimension $k$ and for every pair of indices  $\eta \leq \delta$, and intuitively consist of the homology classes of $\mathbb{X}^{\eta}$ which are still alive at $\mathbb{X}^{\delta}$.

\subsection{Simplicial Complexes}

The second good property of the cover $\mathbb{X}^{\varepsilon}$ is that it can be approximated by a family of simplicial complexes without loosing any topological information. This is crucial because it allows us to work with discrete objects, whose topological invariants can be computed by (means of) simple matrix reduction algorithms. The most intuitive discrete approximation of the cover $\mathbb{X}_n^{\varepsilon}$ is its \emph{Nerve}, also known as Cech complex.\\

\begin{definition}[Cech Complex]
Given a metric space $(\mathbb{X}, d_{\mathbb{X}}) $ the Cech complex $Cech_{\alpha}(\mathbb{X}) $ is the set of simplices $\sigma =[X_1,\ldots,X_k] $ such that the  $k $ closed balls $B(X_i, \alpha)$ have a non empty intersection.
\end{definition} 

Since the elements of $\mathbb{X}^{\varepsilon}_n$ are by definition contractible, $\mathbb{X}^{\varepsilon}_n$ is what is called a \emph{good cover} and it satisfies the assumption of the \emph{Nerve Theorem}.\\

\begin{theorem}[Nerve]
A good cover and its nerve are homotopic.
\end{theorem}

The Nerve Theorem implies that the homology group of $Cech_{\varepsilon} $ are topologically equivalent to those of $\mathbb{X}^{\varepsilon}$. Nevertheless, computing the Cech complex itself can still be computationally challenging; for this reason the Vietoris--Rips complex, another combinatorial representation of $\mathbb{X}_n ^{\varepsilon}$, is typically preferred.\\

\begin{definition}[Vietoris--Rips Complex]
Given a metric space $(\mathbb{X}, d_{\mathbb{X}})$ the Vietoris--Rips complex $Rips_{\alpha}(\mathbb{X})$ is the set of simplices $\sigma = [X_1,\ldots,X_k] $ such that $d_{\mathbb{X}} (X_i, X_j) \leq \alpha$ for all $i,j$.
\end{definition}

The Nerve theorem does not hold for Vietoris--Rips complexes, but its
topology is still close to the one of  $\mathbb{X}_n ^{\varepsilon}$
due to its proximity to the Cech complex:
\[
Rips_{\varepsilon}(\mathbb{X}) \subseteq Cech_{\varepsilon}(\mathbb{X}) \subseteq Rips_{2\varepsilon} (\mathbb{X}).
\]

\section{Persistence Diagrams}

Persistent Homology Groups can be summarized by the
\emph{Persistence Diagram}  $D$, a multiset whose generic element $x_i = (b_i, d_i)$ is the $i^{\tt th}$ generator of a Persistent Homology Group. The first coordinate, the ``birth time'' $b_i$, represent how soon in the filtration the $i^{\tt th}$ feature appears, i.e.~the first value  $\varepsilon$ for which the  $i^{\tt th}$ feature can be found in $\mathbb{X}^{\varepsilon}_n$; the second coordinate, the ``death time'' $d_i$, represents when the feature disappear, i.e.~the first value  $\varepsilon$ for which $\mathbb{X}^{\varepsilon}_n$ does not retain the $i^{\tt th}$ feature anymore. Since two or more feature can share birth and death time, each point has multiplicity equal to the number of features, except for the diagonal, whose points have infinite multiplicity. As death always occurs after birth, all points in the diagram are in or above the diagonal. The \emph{Persistence Barcode} is an equivalent representation, where each bar is a feature whose length correspond to the lifetime of the corresponding feature.

Intuitively the ``lifetime'' or, more formally, the \emph{persistence} $\text{pers}(x) = d - b$, of a feature can be considered as a measure of its importance. Points that are close to the diagonal represent features that appear and disappear almost immediately and may be neglected; a diagram whose only elements are the points of the diagonal $D_{\emptyset}$ is said to be \emph{empty}. In order to distinguish topological signal from topological noise, it possible to build confidence bands around the diagonal \cite{Fasy2014}.

\begin{figure}
\centering
\includegraphics[width = 1\textwidth]{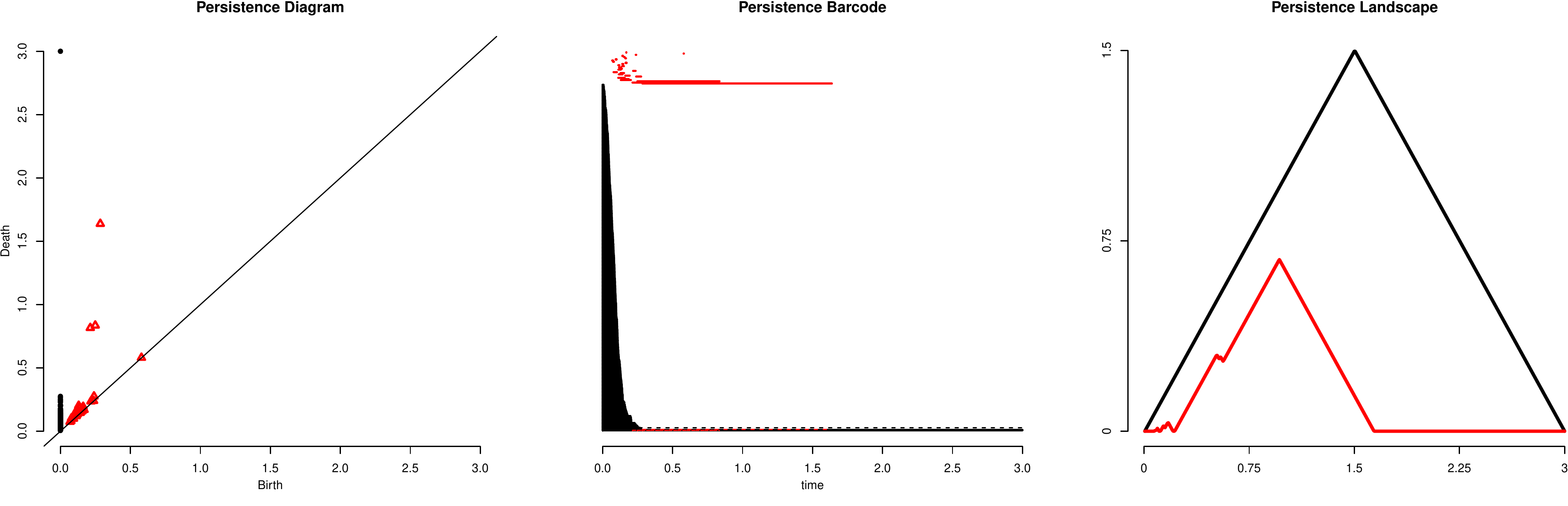}
\caption{(From left to right) Persistence Diagram, Persistence Barcode and Persistence Landscape of data shown in Figure \ref{fig:toydata}.}
\end{figure}

Although in theory a feature may never die, in diagrams built from point clouds all the information is contained between the diagonal and the diameter of the data. For the sake of simplicity, we thus limit our analysis to bounded diagrams.\\

\begin{definition}[Space of Persistence Diagrams] Let $\text{Pers}_p(D) = \sum_{x\in D} \text{pers}(x)^p $ be the degree--$p$ \emph{total persistence} of a persistence diagram $d $. Define the space of persistence diagrams  $\mathcal{D}$ as
\[
\mathcal{D} =  \big\{ D \:\big|\: \text{Pers}_p(D) < \infty \big\},
\] 
where $D_{\emptyset} $ is the persistence diagram containing only the diagonal.
\end{definition}

\subsection{Metrics for Persistence Diagrams}

Persistence diagrams can be compared through several metrics, most noticeably the Bottleneck and the Wasserstein distance, which add to $\mathcal{D}$ the structure of a metric space. The Wasserstein distance, also known as Earth Mover distance or Kantorovich distance is a popular metric in Probability and Computer Science as well as Statistics.

\begin{definition}[Wasserstein Distance between Persistence Diagrams] Given a metric $d$, called \emph{ground distance}, the Wasserstein distance between two persistence diagrams  $D $ and  $D'$ is defined as
\[
W_{d,p} (D, D') =  \left[ \inf_{\gamma} \sum _{x\in D } d\big( x,\gamma (x)\big)^p \right]^{\frac{1}{p}},
\]
where the infimum is taken over all bijections $\gamma : D \mapsto D'$.
\label{def:wass}
\end{definition}

Depending on the choice of the ground distance  $d$, Definition \ref{def:wass} defines a family of metrics, whose most prominent member in \texttt{TDA} literature is the  $L^{\infty}$--Wasserstein distance, $W_{L^{\infty}}$, defined as:
\[
W_{L^{\infty}\!\!, p}(D, D') =  \left[ \inf_{\gamma} \sum _{x\in D } \big\| x-\gamma (x) \big\| _{\infty} ^p \right]^{\frac{1}{p}}.
\]
When $p = \infty$, the distance  $W_{L^{\infty}\!\!, \infty}$ defined as
\[
W_{L^{\infty}\!\!, \infty} (D, D') = \inf_{\gamma} \sup_{x \in D } \big\| x-\gamma (x) \big\|_{\infty},
\]
takes the name of \emph{Bottleneck distance}.

Despite being less popular in the \texttt{TDA} framework, another important choice of ground distance is the $L^2$--norm, especially in the case  $p = 2$, for which \cite{Turner2014} proved that $W_{L^2,2}$ is a geodesic on the space of persistence diagrams.

\begin{proposition}
The space of Persistence Diagrams $\mathcal{D}$ endowed with  $W_{L^2,2} $ is a geodesic space.
\end{proposition}

The space $\mathcal{D}$ is separable and complete in both $W_{L^\infty} $ and $W_{L^2}$, hence is a Polish Space \cite{Mileyko2011}.

\subsection{Stability}

Defining metrics on  $\mathcal{D} $ allows for a notion of stability \cite{Chazal2012}, which, roughly speaking, states that similar topological spaces must have similar diagrams.\\

\begin{theorem}[Stability]
Let  $\mathbb{X,Y} $ totally bounded metric spaces, and let  $D_{\mathbb{X}} $,  $D_{\mathbb{Y}} $ the corresponding persistence diagram, built using either the $Cech$ or the $Rips$ filtration, then
\[
W_{L^{\infty}\!\!, \infty} \left(D_{\mathbb{X}}, D_{\mathbb{Y}} \right) \leq 2 \, d_{\text{GH}} \left( \mathbb{X}, \mathbb{Y} \right),
\]
where $d_{\text{GH}}$ is the Gromov--Hausdorff distance between topological spaces.
\end{theorem} 

Stability is a core result in \texttt{TDA} for two reasons:

\begin{itemize}
\item \textit{the persistence diagram is a topological signature:} stability reassures us that if two point-clouds $\mathbb{X,Y}$ are similar their Persistence Diagrams will be as well, and is therefore instrumental for using them in statistical tasks such as classification or clustering;

\item \textit{the persistence diagram is statistically consistent:} stability reassure us that if we are using a point--cloud $\mathbb{X}_n$ to estimate the topology of an unknown object $\mathbb{X}$, if $\mathbb{X}_n\rightarrow \mathbb{X}$ as $n\rightarrow \infty$, then $D_{\mathbb{X}_n}$ converges to $D_{\mathbb{X}}$ as well.
\end{itemize}

Despite this important property, Persistence diagrams have several drawbacks that have limited their popularity in statistical inference. For example, a collection of Persistence Diagrams  $\{D_1,\ldots,D_n\}$, does not have a unique mean \cite{Turner2014}; moreover despite the fact that $\mathcal{D}$ is a Polish space and that the existence of a probability distribution on it has been proved by \cite{Mileyko2011}, it is still not clear how to derive it. In the following section we will show how kernels can be used to overcome these issues.

\section{Statistical Learning with TDA / Topological Kernels}

In general, the metric structure of the space of persistence diagrams may not be rich enough for statistical learning and, hence we translate them into inner product spaces using kernels. A kernel  $K $ on a space $\mathcal{M}$ is a symmetric binary function $K: \mathcal{M}\times\mathcal{M}\mapsto \mathbb{R}^+$ that can roughly be interpreted as a measure of similarity between two elements of $\mathcal{M}$. Every kernel is associated to an inner product space \cite{Scholkopf2001}; exploiting this correspondence, kernels allow to perform directly most statistical tasks such as classification \cite{Cristianini2000}, regression \cite{Hardle1990}, or testing \cite{Gretton2012}, without explicitly computing, or explicitly knowing, the probability distribution that generated the observations.

\subsection{Geodesic Topological Kernels}

One popular family of kernels for a geodesic metric space $(\mathbb{X}, d)$ is the \emph{exponential kernel}
\[
k(x,y) = \exp\big\{ d(x,y)^p/h \big\} \qquad p,h > 0
\] 
where $h > 0 $ is the bandwidth parameter; for  $p=1 $ this is the Laplacian kernel and for $p=2 $ this is the Gaussian kernel. It is straightforward to use this class to define a \emph{Topological kernel} to be used for statistical
learning.\\

\begin{definition}[Geodesic Topological Kernel] Let  $\mathcal{D} $ be the space of persistence diagrams, and let  $h > 0$, then the Geodesic Gaussian Topological (GGT) kernel $K_{\text{GG}}:\mathcal{D}\times\mathcal{D}\mapsto \mathbb{R^+}$ is defined as
\[
K_{\text{GG}}(D, D') = \exp\left\{\frac{1}{h} W_{L^2,2}(D, D')^2 \right\}  \qquad \forall \, D, D' \in \mathcal{D}.
\]

Analogously, the Geodesic Laplacian Topological Kernel (GLT), $K_{\text{GL}} $ is defined as:\\
\[
K_{\text{GL}}(D, D') = \exp\left\{\frac{1}{h}W_{L^2,2}(D, D') \right\}  \qquad \forall \,  D, D' \in \mathcal{D}.
\]
\end{definition} 

It may seem natural to extend the properties of the standard (Euclidean) Gaussian and Laplacian kernels to their geodesic counterpart on  $\mathcal{D}$, however, it turns out that the metric structure of the space $\mathcal{D}$ may introduce some limitations, especially with respect to positive definiteness; as shown in \cite{Feragen2015}, in fact, a Geodesic Gaussian kernel on a metric space is positive definite only if the space is flat.\\

\begin{theorem}[Feragen et al.]{\label{th:Feragen} }
Let  $(\mathbb{X}, d) $ be a geodesic
metric space and assume that the Geodesic Gaussian kernel on $\mathbb{X}$  $k(x,y) = \exp\{d^2(x,y)/h\}$ is positive definite for all  $h > 0$. Then  $(\mathbb{X}, d)$ is flat in the sense of Alexandrov (see \cite{Bridson1999} for more information).
\end{theorem}

This is not the case for the space of Persistence Diagram, which has been proved to be curved \cite{Turner2014}. We say that a geodesic metric space is $\text{CAT}(k)$ if its curvature is bounded from above by $k$.\\

\begin{theorem}[Turner et al.] {\label{th:Sayan} }
The space of persistence diagrams  $\mathcal{D}$ with $W_{L^2, 2}$ is not  $\text{CAT}(k) $ for any  $k > 0 $, and it is a non--negatively curved Alexandrov space.
\end{theorem}

We can now characterize the Geodesic Gaussian Kernel. 
\begin{lemma} The Geodesic Gaussian Kernel on  $\mathcal{D} $ is not positive definite.
\end{lemma}
The proof is a trivial consequence of Theorem \ref{th:Feragen} and Theorem \ref{th:Sayan}.  Characterizing the Geodesic Laplacian kernel is not as easy, as we show in the appendix, although it has shown empirically to be indefinite as well \cite{Reininghaus2015}. 

\subsection{The competition}

This is not the only, nor the first, attempt to transform persistence diagrams into a more ``inferential--friendly'' object. Previous works in this direction however followed a different strategy and tackled the problem by explicitly deriving a feature map $\Phi:\mathcal{D}\mapsto \mathcal{H} $ from persistence diagrams to some Hilbert space  $\mathcal{H} $. The link between this and our approach is that any feature map  $\Phi $ corresponds to a kernel  $K$ \cite{Cristianini2000, Scholkopf2001} defined as $K(D,D') = \langle\Phi (D),\Phi (D')\rangle_{\mathcal{H}}$, for every $D, D' \in \mathcal{D}$.

We briefly review the two main families of feature maps  $\Phi $: 1. feature maps derived from the Triangle function and 2. feature maps derived from the Dirac Delta function. A common element to the methods presented in the following is that the embedding is defined point--wise, for each element of the persistence diagram, at first. The structure of the diagram must be later recovered as a summary, whereas the geodesic kernel maintains it directly, as it always consider the the persistence diagram as a whole.

\paragraph{Triangle Function} The first way of translating each point $x \in D $ into a space of function is through the triangle function $T_x(t)$ defined as 
\[
T_x (t) =
 \begin{cases}
t-b+d & t\in [b-d,b], \\
b+d-t & t\in (b, b+d], \\
0 & \text{otherwise}.
\end{cases}
\]
Informally, this corresponds to linking each point of the
diagram to the diagonal with segments parallel to the axis and then rotating the diagram of $45$ degrees; each diagram is then represented by a collection of piecewise linear functions $\{T_x\}_{x\in D} $. Persistence Landscapes  $\lambda_{D} (k,t)$ \cite{Bubenik2015} are defined by taking the $k^{\tt th}$ outermost line of the collection, i.e.
\[
\lambda_{D} (k,t) = k\text{-}\!\max _{x\in D} T_x(t) \qquad k \in \mathbb{N}^+,
\]
where  $k\text{-}\!\max$ is the $k^{\tt th}$ largest value in the set $T_x (t)$. It immediately follows that for any given $k\in\mathbb{N}^+ $ the feature map  $\Phi(D)$ is defined as $t\mapsto \lambda_{D} (k,t)$. Persistence Silhouettes  $\psi(t)^q$ \cite{Chazal2014a} are defined as weighted average of the triangle functions, i.e.
\[
\psi(t) ^q = \frac{\sum_{x \in D} \text{pers}(x)^q \, T_x(t)}{\sum_{x\in D}\text{pers}(x)^q} \qquad q \in \mathbb{R}^+.
\]

In theory it is possible to define a kernel from the Persistence Landscape  $K_{\lambda}(D,D') $ (and analogously for the Silhouette), but since in practice it has shown poor performances (as shown in \cite{Reininghaus2015}), these tools are typically used as they are or summarized in some other way.

Persistence Landscapes and Persistence Silhouettes are both defined in Hilbert spaces, they can be averaged, they allow for a Law of Large Numbers and Central Limit Theorem and they can be \emph{robustified} through sub--sampling \cite{Chazal2015}, while still retaining topological information. However nor the Persistence Landscape, nor the Persistence Silhouette, are injective: it is not possible to transform Landscapes back to Persistence Diagrams, and thus the interpretation of the average Landscape (or analogously the average Silhouette) is often challenging.

\paragraph{Dirac Delta Functions} The second way of mapping each $x \in D$ to a space of function is through Dirac delta functions $\delta_x$. Every Persistence diagram  $D$ can be uniquely represented as the sum of Dirac delta functions  $\delta_{x}$, one for each  $x \in D$; since  $\delta_x$ are defined in a Hilbert space, their sum will as well.

Reininghaus et al. (2015) use this representation as initial condition for a heat diffusion problem, and define a new feature map  $\Phi(D)$ as
\[
t\mapsto \frac{1}{4\pi\sigma}\sum_{x\in D} \mathtt{e}^{-\frac{\norm{ t-x} ^2}{4\sigma}} - \mathtt{e}^{-\frac{\norm{ t-\bar{x}}^2}{4\sigma}},
\]
where if $x=(b, d) $ then  $\bar{x} = (d,b) $. The feature map $\Phi(D) $ defines the Persistence Scale Space kernel $K_{\text{PSS}}$:
\[
K_{\text{PSS}}(D, D') = \frac{1}{8\pi\sigma}\sum_{x \in D} \sum_{y \in D'} \mathtt{e}^{-\frac{\norm{ x-y }^2}{8\sigma}} - \mathtt{e}^{-\frac{\norm{ x-\bar{y} }^2}{8\sigma}} \qquad \forall \, D, D' \in \mathcal{D},
\]
which is the most similar in spirit to the Geodesic Kernels. $K_{\text{PSS}}$ is a heat kernel, and is stable with respect to $W_{L^{\infty}\!\!,1}$.

Another kernel built from Dirac Delta functions is the Persistence Weighted Gaussian Kernel \cite{Kusano2016}, defined as 
\[
K_{\text{PWG}}(D,D') = \exp\left(-\frac{d_{\text{G}}(D,D')^2}{2 \, \sigma^2}\right)
\]
where
\[ 
\begin{split}
d_{\text{G}}(D, D')  & =\sum_{x\in D}\sum_{x' \in D} w_{\text{arc}}(x) \, w_{\text{arc}}(x') \, k_{\text{G}}(x,x')\\
 & +\sum_{y\in D'}\sum_{y' \in D'} w_{\text{arc}}(y) \, w_{\text{arc}}(y') \, k_{\text{G}}(y,y') \\
 & -2\sum_{x\in D}\sum_{y \in D'} w_{\text{arc}}(x) \, w_{\text{arc}}(y) \, k_{\text{G}}(x,y), \\
 & \\
w_{\text{arc}}(x)& =\arctan \big( C \cdot \text{pers}(x)^q \big), \\
\end{split}\]

and $k_{\text{G}}$ is the Euclidean Gaussian kernel with variance  $\tau$. The Persistence Weighted Gaussian Kernel, much like the Persistence Silhouette, allows to explicitly control the effect of persistence.
However, the choice of the different  $4 $ tuning parameters ($q, \sigma, \tau, C$) may be unfeasible in most real data applications.

The main difference with respect to our Geodesic Kernels is that $K_{\text{PSS}}$,  $K_{\text{PWG}}$ and even $K_{\lambda} $ are positive definite by construction. Despite being indefinite, however, the Geodesic Kernels are a more sensible measure of similarity.

\begin{figure}
\includegraphics[width = 1\textwidth]{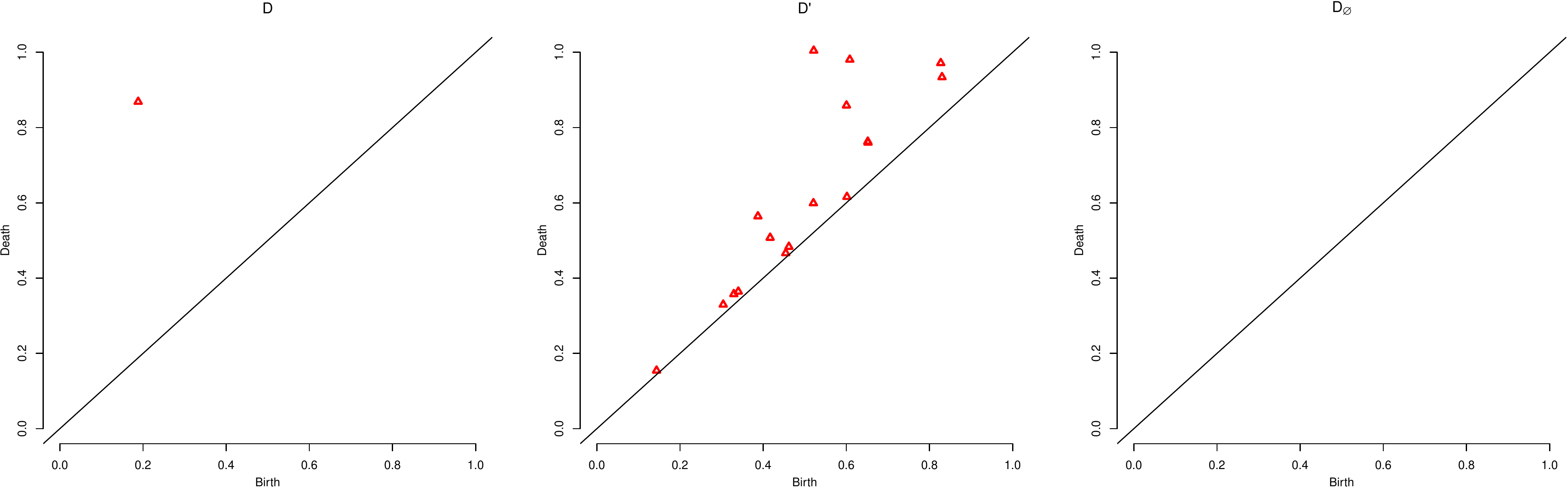} 
\caption{(From left to right) Three Persistence diagrams: $D$, $D'$, $D_{\emptyset}$.}
\label{fig:foodiag}
\end{figure}
\begin{table}
\centering

\begin{tabular}[t]{c|rrr}
& $D$ & $D'$ & $D_{\emptyset}$\\
%\hline
\hline
$D$ & 1.000 & 0.040 & 0.483\\
%\hline
$D'$ & 0.040 & 1.000 & 0.006\\
%\hline
$D_{\emptyset}$ & 0.483 & 0.006 & 1.000\\
%\hline
\end{tabular}
\caption{Geodesic Gaussian Kernel matrix for the three diagrams shown in Figure \ref{fig:foodiag}.}
\label{tab:geok}

\begin{tabular}[t]{c|rrr}
& $D$ & $D'$ & $D_{\emptyset}$\\
\hline
%\hline
$D$ & 0.005 & 0.023 & 0.000\\
%\hline
$D'$ & 0.023 & 0.119 & 0.000\\
%\hline
$D_{\emptyset}$ & 0.000 & 0.000 & 0.000\\
%\hline
\end{tabular}
\caption{Persistence Scale Space Kernel matrix for the three diagrams shown in Figure \ref{fig:foodiag}.}
\label{tab:pssk}

\end{table}

Let us examine the behavior of the kernels with respect to the empty diagram  $D_\emptyset$ to make this more clear. This will be especially relevant later, when analysing posturography data (see Section \ref{sec:class}). Although not all diagrams are equally different from the empty diagram  $D_\emptyset$, $K_{\text{PSS}}$ and $K_{\text{PWG}}$ do not capture this diversity as neatly as the Geodesic Kernels.

In the  $\text{PSS}$ approach, for example, $\Phi(D_{\emptyset}) = 0$ by definition. This results in $K_{\text{PSS}}(D_{\emptyset},D) =\langle \Phi(D_{\emptyset}), \Phi(D) \rangle = 0$, for every  $D\in \mathcal{D} $, including  $D_\emptyset$ itself, leading to the paradoxical conclusion that $K_{\text{PSS}}(D_{\emptyset},D_{\emptyset}) = 0$, as shown in Table \ref{tab:pssk}.

The Geodesic Kernels, on the other hand, are built on the Wasserstein distance and since  $W_{L,p}(D,D_\emptyset) \neq 0 $ for any $D \neq D_\emptyset$, they retain more information, as can be seen in Table \ref{tab:geok}.

Although positive definiteness is a rather attractive quality in a kernel \cite{Scholkopf2001}, the indefiniteness of our kernel does not affect its performances in supervised settings. Notice that we are not claiming that our kernel is superior, in fact due to their positive definiteness $K_{\text{PSS}}$ and $K_{\text{PGW}}$ are far more general and can be used in any kernel algorithm. We are instead proposing an alternative that exploiting the predictive power of the negative part of the kernels can perform better in a narrower class of problems. We now show some applications to real data to support our thesis.

\section{Regression / Fullerenes}

Buckyballs fullerenes are spherical pure carbon molecules artificially synthesized in the '$70$, then discovered in nature in the '$90$, which have recently gained much attention after \textsf{C60} has being identified as the largest molecule detected in space \cite{Berne2012}. The typical trait of Buckyballs fullerenes is that atoms' linkage can form either pentagons or hexagons, so that the configuration of the molecule resembles a soccer ball (hence the name). Our goal is to show that the topology of the molecule can be used directly to explain its Total Strain Energy (measured in  $Ev $); given a sample $\{X_1,\ldots, X_n\}$ of Fullerenes we model their Total Strain Energy, $Y$ as a function of their Persistence Diagrams $\{D_1,\ldots, D_n\}$: 
\[
Y_i = m(D_i) + \varepsilon_i \qquad \forall \, i \in \{1,\ldots, n\},
\] 
where $\varepsilon_i$ is the usual  $0$--mean random error. 

As in standard nonparametric regression, we can estimate the regression function  $m(\cdot) $ with the Nadaraya--Watson estimator \cite{Hardle2012}, defined as:
\[
\widehat{m}(D) =\frac{\sum_{i=1}^n Y_i \, K(D, D_i)}{\sum_{i=1}^n K(D, D_i)},
\]
where $D$ is a generic persistence diagram. Since the kernel function  $K $ involved in the Nadaraya--Watson estimator, needs not be positive definite, we can use the Geodesic kernels to extend nonparametric regression to the case of persistence diagrams as covariate.

\begin{figure}
 \centering
    \includegraphics[width=0.19\textwidth]{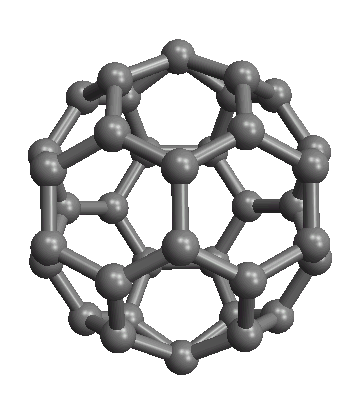}
    \includegraphics[width=0.19\textwidth]{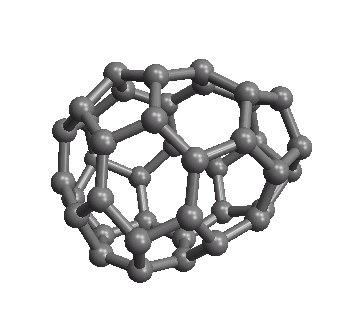}
    \includegraphics[width=0.19\textwidth]{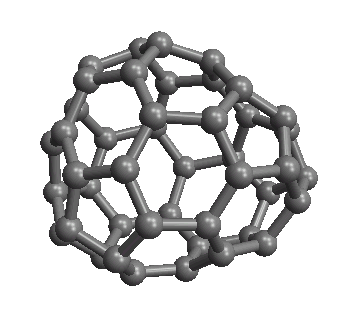}
    \includegraphics[width=0.19\textwidth]{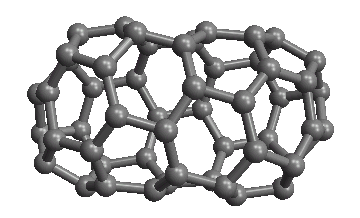}
    \includegraphics[width=0.19\textwidth]{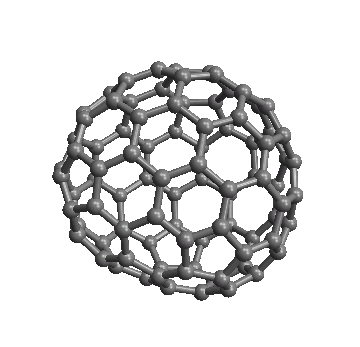}
    \includegraphics[width=1\textwidth]{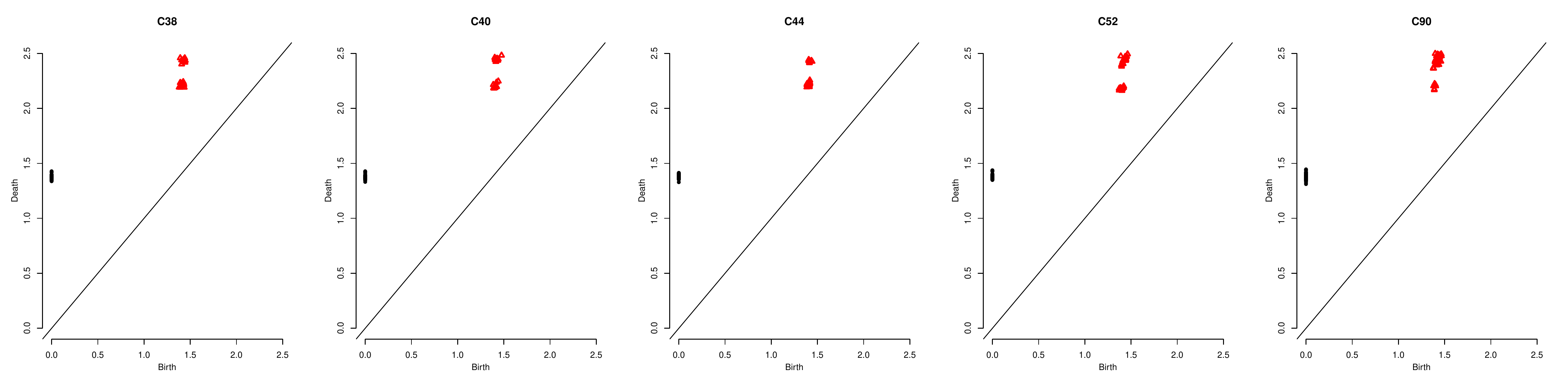}
    \caption{Topological configurations of some fullerenes (top) and corresponding persistence diagrams (bottom). From left to right: \textsf{C38}(\textsf{C2}v), \textsf{C40}(\textsf{C1}), \textsf{C44}(\textsf{C1}), \textsf{C52}(\textsf{C2}), \textsf{C90}(\textsf{C1}).}
    \label{fig:descriptive}

 \centering
    \includegraphics[width=1\textwidth]{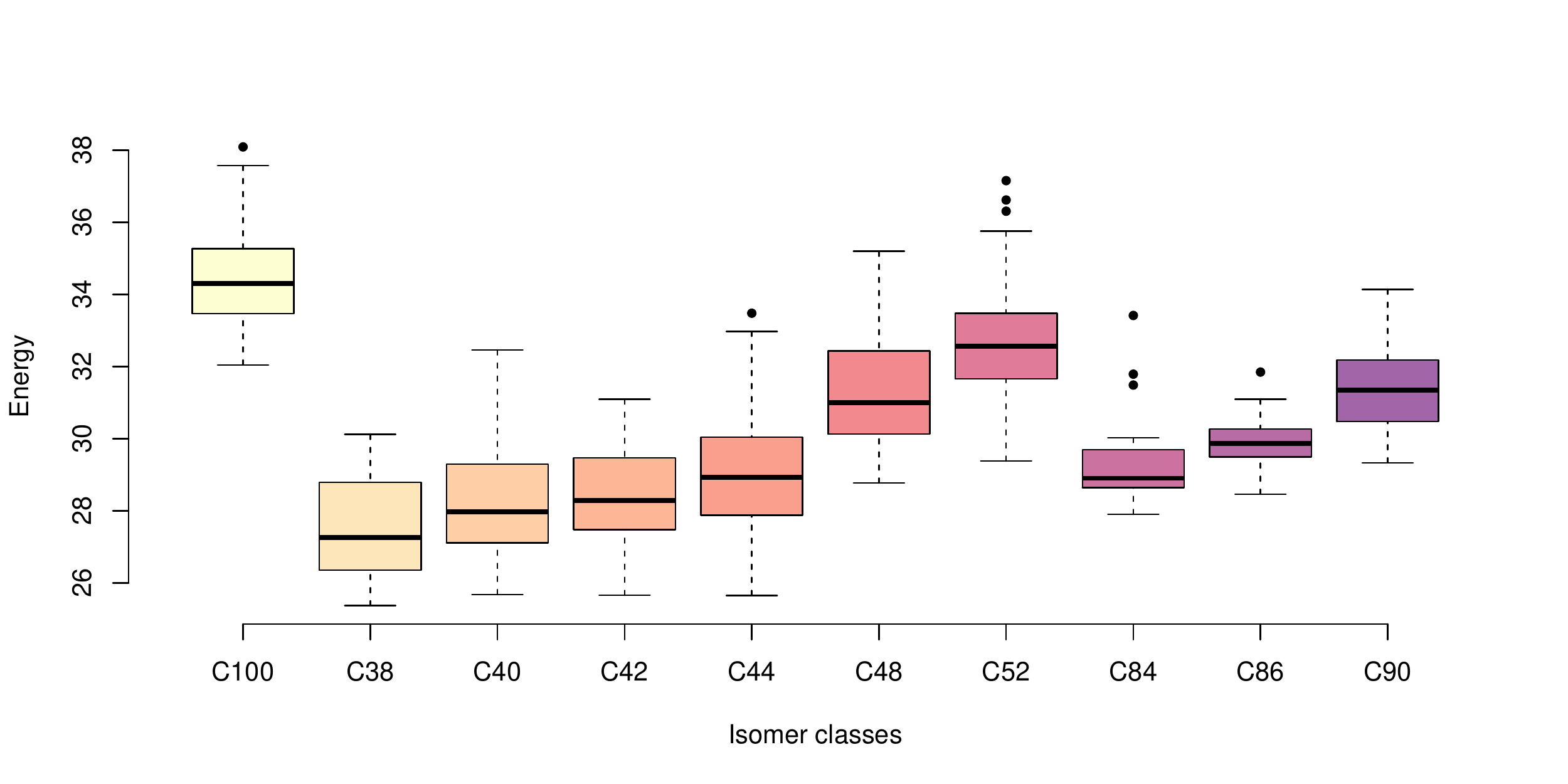}
    \caption{Energies for the 10 different classes of isomers. It is worth noticing that Fullerenes with higher numbers of atoms do not necessarily have higher energy.}
    \label{fig:boxplot}
\end{figure}
We fit the model using data from  $n = 535$ molecules of  $10$ different types of Fullerenes. The sample is unbalanced, as the number of configurations available for each Fullerene depends on the number of atoms composing it and andvances in research (Table \ref{tab:fullerenes}).

\begin{table}
\centering
\begin{tabular}{l|rrrrrrrrrr}
  & \textsf{C38} & \textsf{C40} & \textsf{C42} & \textsf{C44} & \textsf{C48} & \textsf{C52} & \textsf{C84} & \textsf{C86} & \textsf{C90} & \textsf{C100} \\
\hline
%\hline

$n$  & 17 & 40 & 45 & 89 & 79 & 96 & 24 & 19 & 46 & 80\\

$\bar{Y}$  & 27.50 & 28.29 & 28.46 & 29.12 & 31.21 & 32.59 & 29.34 & 29.88 & 31.29 & 34.41\\

$\widehat{\sigma}$  & 1.35 & 1.62 & 1.35 & 1.78 & 1.56 & 1.57 & 1.29 & 0.80 & 1.21 & 1.24\\
%\hline
\end{tabular}
\caption{Number of observations ($n$), mean ($\bar{Y}$) and standard deviation ($\widehat{\sigma}$) of TSE for each type of fullerenes in the sample.}
\label{tab:fullerenes}
\end{table}

\begin{figure}
 \centering
    \includegraphics[width= 0.9\textwidth]{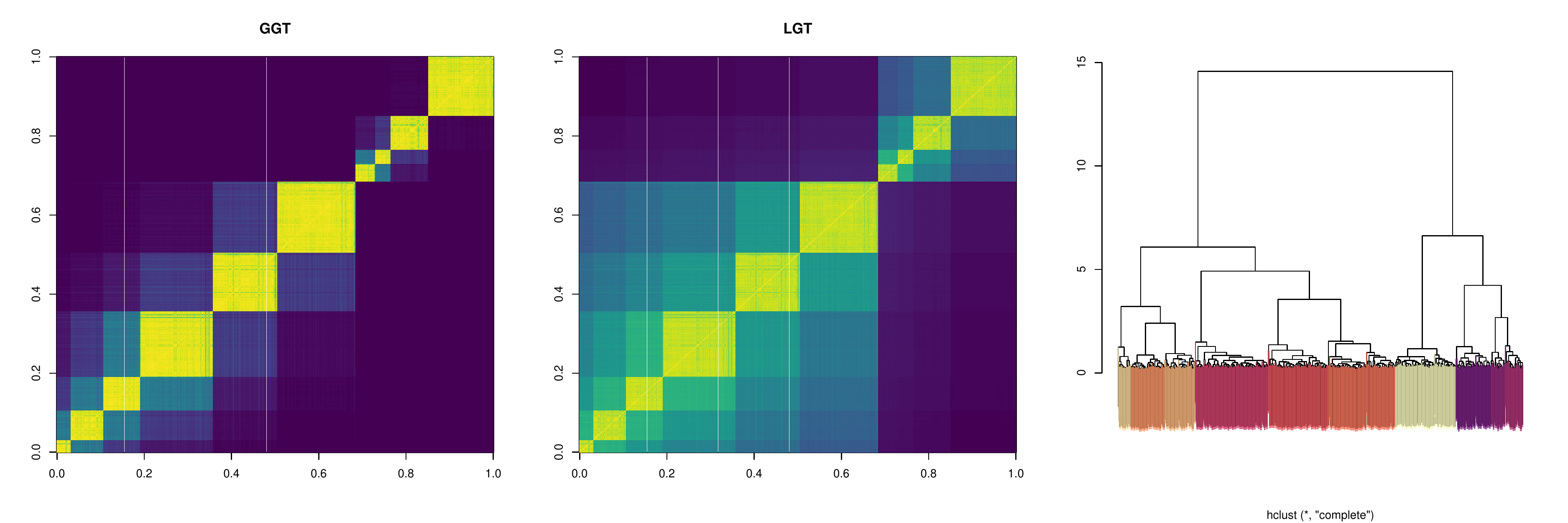}
    \caption{Kernel Matrix for the Geodesic Gaussian Kernel (left), Geodesic Laplacian Kernel (center), Hierarchical Clustering built from the Wasserstein distance with complete linkage (right). Colors represent the different isomer classes as shown in Figure \ref{fig:boxplot}.}
    \label{fig:kernmat}
\end{figure}

For each molecule, the data (freely available at \url{http://www.nanotube.msu.edu/fullerene/fullerene-isomers.html} consists of the coordinates of the atoms taken from Yoshida's Fullerene Library and then re--optimized with a Dreiding--like forcefield. 
We carry our analysis using both the \texttt{R} package \texttt{TDA} \cite{Fasy2014b} and the \texttt{C++} library it refers to \href{http://www.mrzv.org/software/dionysus}{\tt Dionysus}  \cite{Morozov2012}. 
We started by building the persistence diagrams encoding data into the Rips filtration. Since there is no clear pattern for connected components and, as we could expect, there is only one relevant void for each molecule, we decided to focus on features of dimension $1$, which seem to be the most informative. As we can see from Figure \ref{fig:descriptive}, loops in the diagrams are, in fact, clearly clustered around two centers, which represent the pentagons and the hexagons formed by the carbon atoms. Interestingly enough, the Wasserstein distance and, hence, both the geodesic kernels, fully recover the class structure induced by the isomers, as we can see in Figure \ref{fig:kernmat}.

We estimate the regression function $m(D)$ using both the Laplacian and the Gaussian geodesic kernels; the estimator resulting from the \text{GGT} kernel is
\[
\widehat{m}_{\text{GG}}(D) =\frac{\sum_{i=1}^n Y_i \exp\left\{-\frac{1}{h} W_{L^\infty\!\!,2} (D, D_i)^2 \right\} }{\sum_{i=1}^n \exp\left\{-\frac{1}{h} W_{L^\infty\!\!,2} (D, D_i)^2 \right\}} \qquad \forall \, D \in \mathcal{D};
\]
analogously for the \text{LGT} kernel.

\begin{table}
\centering
\begin{tabular}{l|rr}
& Geodesic Gaussian Kernel & Geodesic Laplacian Kernel \\
\hline
%\hline
Nonparametric regression  & $339.89$  & $342.14$ \\ 
Semiparametric regression & $1049.02$ & $331.04$ \\
%\hline
\end{tabular}
\caption{Residual Sum of Squares.}
\label{tab:fullerenesRSS}
\end{table}

Moreover, in order to take into account the group structure naturally
induced by the isomers, we considered a model with a fixed group
intercept, i.e: 
\[
Y_{ij} = \alpha_j + m(D_{ij}) + \varepsilon_{ij},
\]
where  $D_{ij} $ denotes the persistence diagram of the  $i^{\tt th}$ isomer of the $j^{\tt th}$ molecule. We fit the resulting partially linear model using Robinson's trimmed estimator, as detailed in \cite{Li2007}.

After choosing the bandwidth  $h$ via Leave--One--Out cross validation, we compare the different models in terms of Residual Sum of Squares (RSS). As we can see from Table \ref{tab:fullerenesRSS}, the two kernels yield similar results when used in a fully nonparametric estimator, while the Laplacian kernel performs better when adding the group intercept to the model. This can be understood by looking at the kernel matrices (Figure \ref{fig:kernmat}); the Gaussian Kernel has a sharper block structure than the Laplace Kernel, which makes it better at discriminating the $10$ molecule classes. However, when the group structure is taken into account by the model itself, this clustered structure leads to worse prediction.
\begin{figure}
 \centering
    \includegraphics[width=1\textwidth]{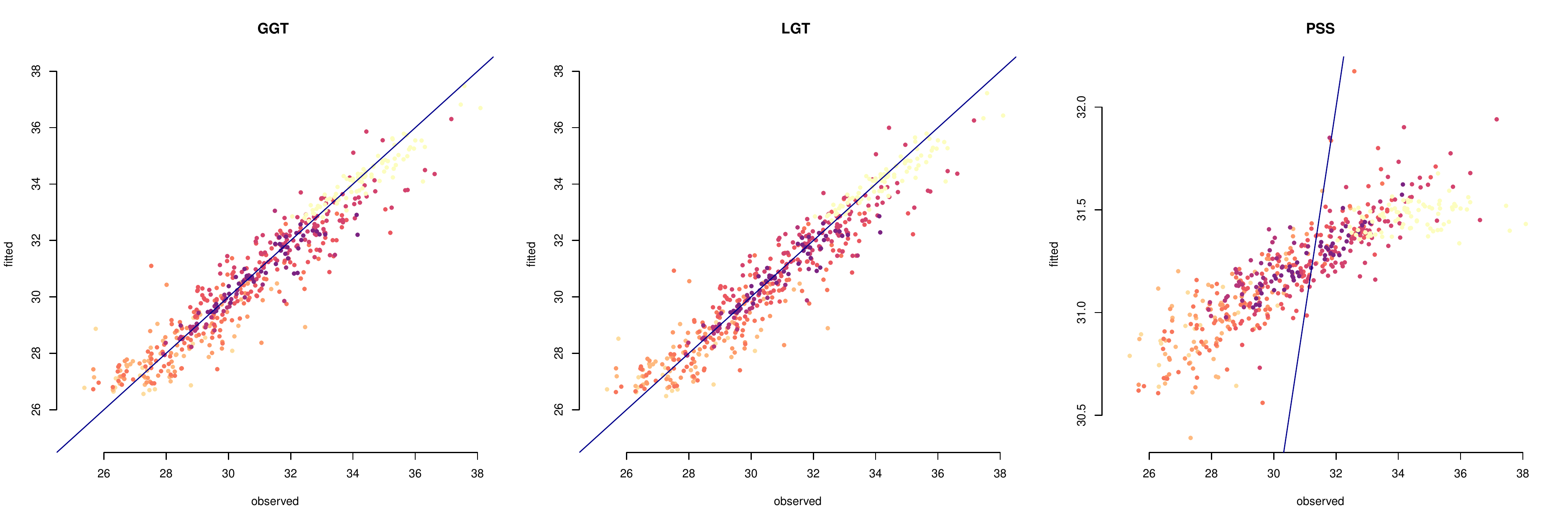}
    \caption{Observed vs fitted plot for the fully nonparametric model fitted with the Geodesic Gaussian (left), Geodesic Laplacian (center) and the Persistence Scale Space kernel (right). Colors represent the different isomer classes as shown in Figure \ref{fig:boxplot}.}
    \label{fig:fitvsobs}
\end{figure}

Finally, we compare the performance of our geodesic kernels with the Persistence Scale Space kernel  $K_{\text{PSS}}$ by using the same data to fit
\[
\widehat{m}_{\text{PSS}}(D) =\frac{\sum_{i=1}^n Y_i \, K_{\text{PSS}}(D, D_i)}{\sum_{i=1}^n K_{\text{PSS}}(D, D_i)}.
\]
As we can clearly see from the fitted-vs-observed plots in Figure \ref{fig:fitvsobs}, the positive definiteness of the  $\text{PSS}$ kernel does not result in more accurate prediction, as both $K_{\text{GG}}$ and $K_{\text{LG}}$ outperform it.

\section{Classification / Posturography}\label{sec:class}

For our second example we analyze data from a posturography experiment available at \url{https://physionet.org/physiobank/database/hbedb/}. Subjects standing on a platform were asked to close their eyes and stand still for some time. Researcher then recorded the center of pressure on the platform over a period of  $60$ seconds; details are available in \cite{Santos2016}. In order to characterize the oscillation's pattern using \texttt{TDA}, we build a Rips diagram for each of the  $320$ traces,  $160$ of which were recorded on a rigid platform, and  $160$ on a soft one. 

\begin{figure}[ht]
 \centering
    \includegraphics[width=.65\textwidth]{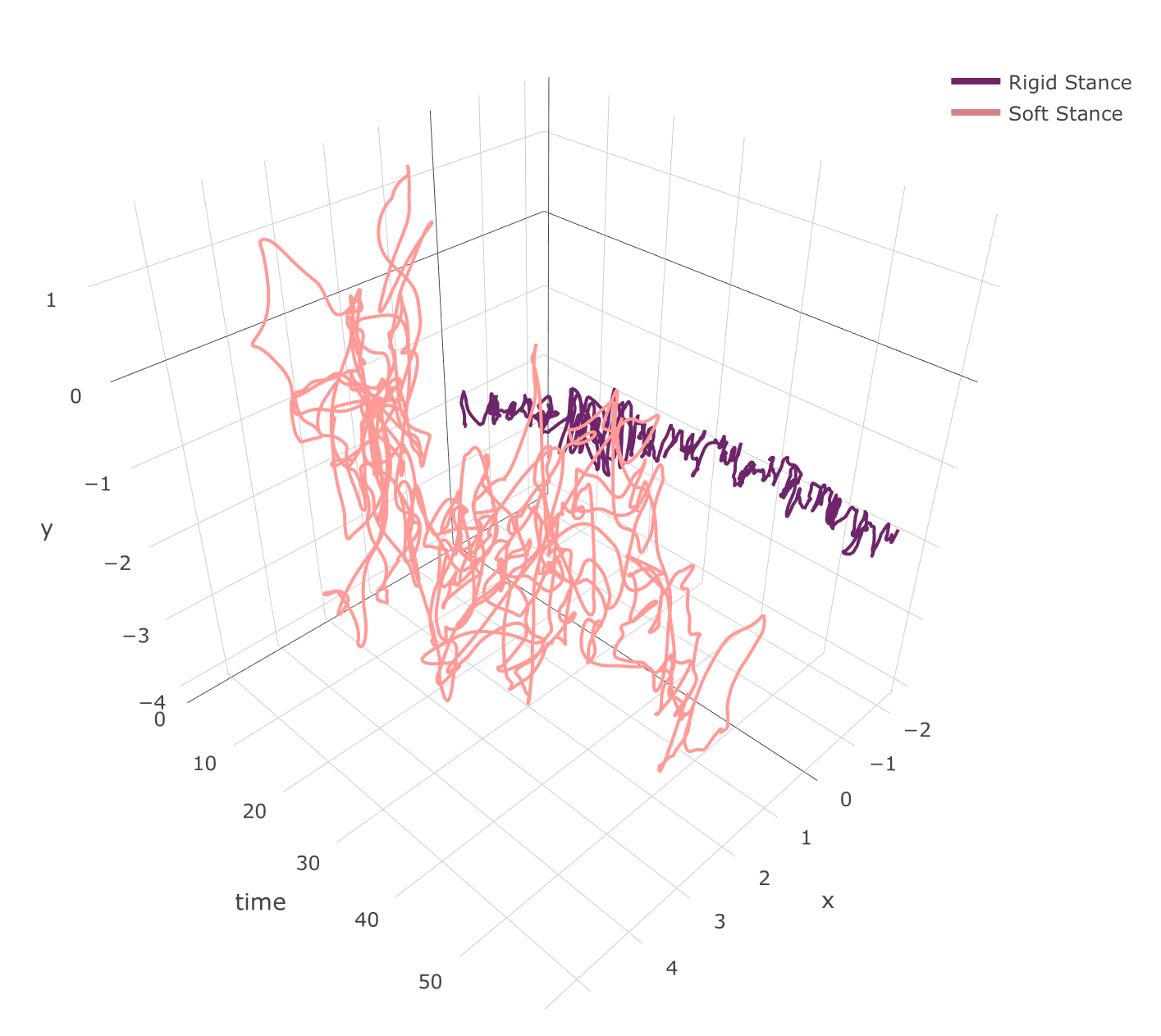}
    \includegraphics[width=.3\textwidth]{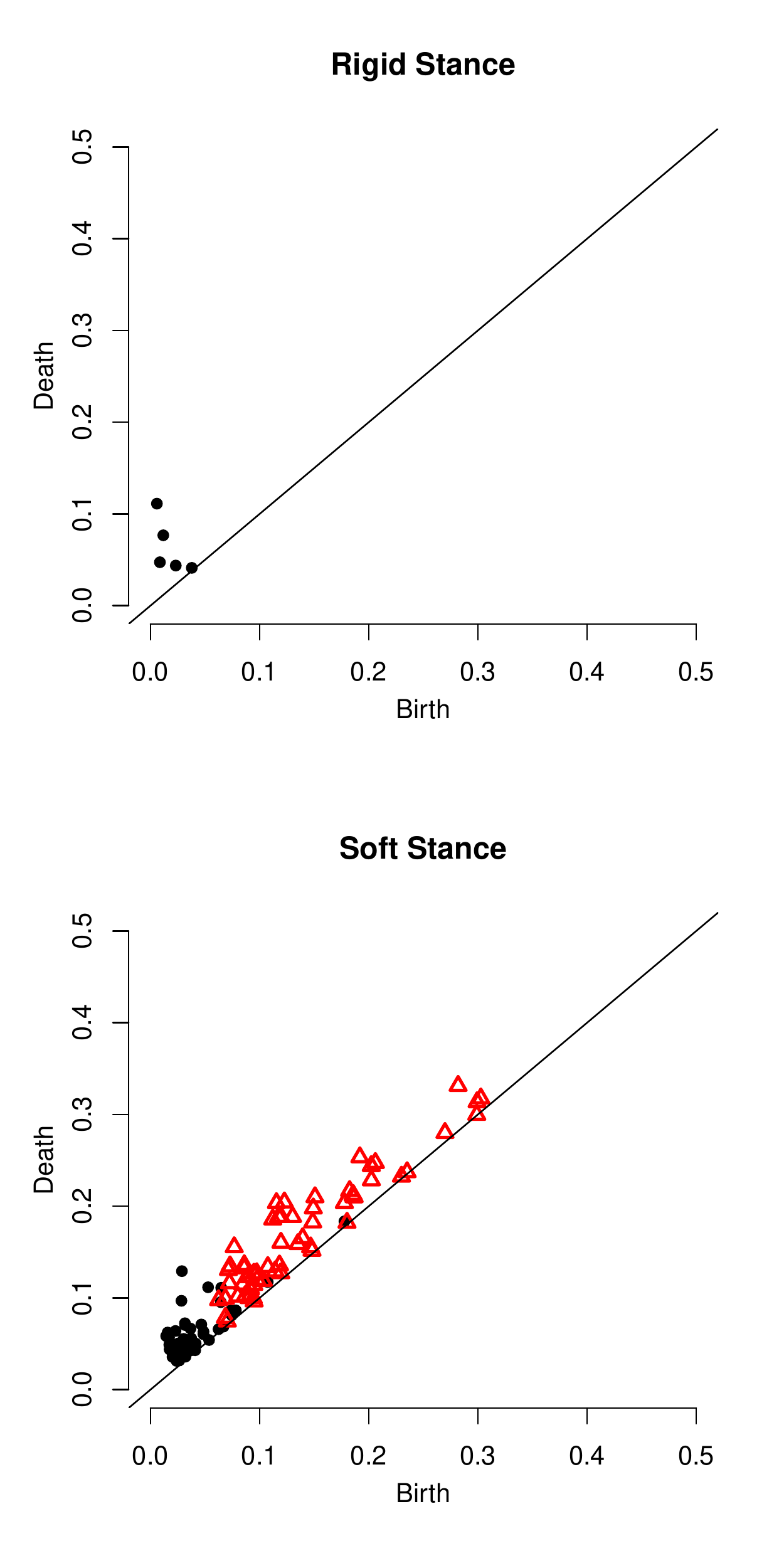}
    \caption{Trajectory of a subject standing on a soft platform (in pink) and on a rigid one (in brown). On the left, the corresponding persistence diagrams. }
    \label{fig:post}
\end{figure}

We focus on dimension $1$ topological features. Intuitively, in fact, a loss of equilibrium results in sudden movements, which generate cyclical structures; we can consider the number and the persistence of loops as a measure of the signal's variability. Figure \ref{fig:post} shows one trajectory for each of the conditions. Data coming from the rigid platform do not present any loop at all, causing the diagram to be empty (as we are only considering dimension $1$ features).

Although not all observations are quite as well distinguishable, it is generally true that subjects standing on the rigid platform are more stable and their persistence diagrams are more likely to be empty. 

This kind of data fits perfectly in
the \texttt{TDA} framework, as coordinates, which in this case represent
the direction and the time of the loss of equilibrium, are not relevant
to our problem and may be misinterpreted.

We show how the Persistence Diagram of a trace can be used to infer whether each trajectory was recorded on a soft or a rigid platform. This is a binary classification problem, which we solve using the Geodesic Kernels. Standard kernel--based classifiers such as Support Vector Machines require a positive definite kernel, we thus consider an extension to SVM for indefinite kernels proposed by \cite{Loosli2016}, KSVM. Details are given in the Appendix.

As we can see from Table \ref{tab:miscl}, the accuracy of the classification is far superior when using \text{KSVM} with the Geodesic Gaussian Kernel  $K_{\text{GG}}$ (and results are identical for  $K_{\text{LG}}$) rather than the standard SVM with the positive definite  $K_{\text{PSS}}$, and this result is not surprising because several of the diagrams corresponding to trajectories on the Rigid Platform are empty. 

\begin{table}[ht]
\centering
\begin{tabular}[t]{l|rrrrr}
&KSVM & PSS--SVM & Clip & Flip & Square\\
\hline
%\hline
Mean       & $2.82\%$ & $3.31\%$  & $2.84\%$ &$2.86\%$ & $2.87\%$\\
Standard Deviation &  $0.087$ & $0.198$ & $0.159$ & $0.141$ & $0.166$\\
%\hline
\end{tabular}

\caption{Average Misclassification Rate for the $10$--fold Cross Validation and corresponding variance.}
\label{tab:miscl}
\end{table}
Although there are algorithms, such as KSVM and others \cite{Luss2008}, designed to explicitly solve the SVM optimization problem when the kernel is indefinite, a very common way to deal with indefinite kernels $K$ is to just substitute the kernel matrix $\mathbb{K}$, whose $(i,j)^{\tt th}$ entry is defined as $\mathbb{K}_{ij} = K(D_i, D_j)$, with some positive definite approximation of it. Denote by $\mathbb{K} = U \, \Lambda \, U^{\tt t}$ the spectral decomposition of the indefinite matrix $\mathbb{K}$, where $U$ is an orthogonal matrix and $\Lambda = \mathrm{diag}(\lambda_1,\ldots,\lambda_n)$ is the diagonal matrix of (real by symmetry) eigenvalues $\{\lambda_1, \ldots, \lambda_n\}$. We consider the following heuristics to obtain a positive definite kernel matrix $\widetilde{\mathbb{K}}$:
\begin{itemize}
\item \textbf{clip:} set to $0$ negative eigenvalues of $\mathbb{K}$; that is, $\widetilde{\mathbb{K}}_{\tt c} = U\,\widetilde{\Lambda}_{\tt c} \,U^{\tt t}$ where 
\[
\widetilde{\Lambda}_{\tt c} = \mathrm{diag}\big(\max(\lambda_1,0),\ldots,\max(\lambda_n,0)\big);
\]
\item \textbf{flip:} take the absolute value of the eigenvalues of $\mathbb{K}$; that is, $\widetilde{\mathbb{K}}_{\tt f} = U\,\widetilde{\Lambda}_{\tt f} \,U^{\tt t}$ where 
\[
\widetilde{\Lambda}_{\tt f} = \mathrm{diag}\big(|\lambda_1|,\ldots,|\lambda_n|\big);
\]
\item \textbf{square:} square the eigenvalues of $\mathbb{K}$; that is, $\widetilde{\mathbb{K}}_{\tt s} = U\,\widetilde{\Lambda}_{\tt s} \,U^{\tt t}$ where
\[
\widetilde{\Lambda}_{\tt s} = \mathrm{diag}\big(\lambda_1^2,\ldots,\lambda_n^2\big).
\]
\end{itemize}

We compare the performance of KSVM with that of a standard SVM trained on $\widetilde{\mathbb{K}}$. 
The three heuristics we consider in order a positive definite version of the kernel matrix  $\mathbb{K} $. Results in Table \ref{tab:miscl} are rather reassuring, since they suggest that the good performance of the KSVM with $K_{\text{GG}} $ it does not depend on the complexity of the specific solver, but rather on the discriminative power of the Geodesic Kernels themselves.

\subsubsection*{Conclusions}

Topological Data Analysis is an exciting new field that has seen a
tremendous growth in the last couple of years. The theoretical
developments have, however, not been matched with popularity in
applications, as topological summaries are defined in rather complex
spaces. In contrast with most of the \texttt{TDA} literature we thus presented a
practical framework for this new set of tools. We defined a new class of
kernels, the Geodesic Topological kernels, which retains more
information than other previously defined kernels, and we showed how to
exploit them in the context of supervised learning, where their
indefiniteness can be easily overcome. In this limited setting, our kernel has shown promising behavior, as it outperformed other topological kernels; we now plan to extend our investigation to the unsupervised setting. 

Finally, results presented here are encouraging for the emerging branch of \texttt{TDA} in general. To the best of our knowledge, this is, in fact, the first time that persistence diagrams are used as covariates directly and highlights the potential of \texttt{TDA} in yet another setting.  

\medskip
\bibliographystyle{siam}
\bibliography{bibbiKernels}

\newpage
\appendix
\subsubsection*{Appendix -- RKKS}

Given a dataset $\mathcal{D}_n = \{(x_i, y_i)\}_{i=1}^n$, in its standard formulation in a Reproducing Kernel Hilbert Space $\mathcal{H}$ -- i.e. a space generated by a positive definite kernel $K$ -- Support Vector Machine (SVM) is defined as the solution to following optimization problem: 
\begin{equation}
\begin{cases} \underset{f\in \mathcal{H}, b\in \mathbb{R}}{\min} \!\!\! & \frac{1}{2}\norm{ f} ^2 _{\mathcal{H}}  = \frac{1}{2} \langle f,f \rangle_{\mathcal{H}}, \\
\text{s.t} & \sum_{i=1}^n \max\big\{ 0, 1 - y_i \, \big( f(x_i) + b \big) \big\} \leq \tau,
\end{cases}
\label{eq:SVM}
\end{equation}
or equivalently, in its dual form: 
\[
\begin{cases}
\underset{\boldsymbol{\alpha}}{\max} \!\!\! & -\frac{1}{2} \, \boldsymbol{\alpha}^{\tt t}  \mathbb{G} \, \boldsymbol{\alpha} + \boldsymbol{\alpha}^{\tt t} \boldsymbol{1} - \mu \, \boldsymbol{\alpha}^{\tt t}  \boldsymbol{y}, \\
\text{and} & \big| \alpha_i \big| \leq \eta, \quad i \in \{1,\ldots,n\},
\end{cases}
\] 
where $\boldsymbol{1} \in \mathbb{R}^n$ is a vector of all ones,  $\eta$ is the slack variable and  $\mathbb{G}$ the kernel matrix such that  $\mathbb{G}_{ij} = y_i \, y_j \, k(x_i,x_j)$.

Extending to the indefinite kernels standard kernel--based classifiers such as Support Vector Machine (SVM) requires some knowledge about Reproducing Kernel Krein Spaces \cite{Alpay1991, Gheondea2013}. Every positive kernels are associated to RKHS, similarly each indefinite kernel is associated to a Reproducing  Kernel Krein Space (RKKS). A RKKS $\mathcal{K}$ is an inner product space endowed with a Hilbertian topology for which there are two RKHS $\mathcal{K}_{+}$ and  $\mathcal{K}_{-}$ such that
\[
\mathcal{K} = \mathcal{K}_{+} \oplus \mathcal{K}_{-}.
\]
RKKS share many properties of RKHS, most noticeably the Riesz and the Representer theorem, which allow to define a solver for the SVM problem.

It has been proven, \cite{Ong2004}, that a minimization problem in a RKHS can be translated into a \emph{stabilization problem} in a RKKS. The SVM optimization problem in a RKKS  $\mathcal{K} $ thus can be written as:
\[
\begin{cases}
\underset{f\in \mathcal{K}, b\in \mathbb{R}}{\text{stab}} \!\!\! & \frac{1}{2} \langle f,f \rangle_{\mathcal{K}} \\
\text{s.t} & \sum_{i=1}^n \max \big\{ 0, 1 - y_i \, \big( f(x_i) + b \big)\big\} \leq \tau,
\end{cases}
\] 
which \cite{Loosli2016} proved that can also be written in its dual form

\begin{equation}
\begin{cases}
\underset{\widetilde{\boldsymbol{\alpha}}}{\max} \!\!\! & -\frac{1}{2} \, \widetilde{\boldsymbol{\alpha}}^{\tt t}  \widetilde{\mathbb{G}} \, \widetilde{\boldsymbol{\alpha}} + \widetilde{\boldsymbol{\alpha}}^{\tt t} \boldsymbol{1} - \mu \, \widetilde{\boldsymbol{\alpha}}^{\tt t}  \boldsymbol{y}, \\ 
$\text{and}$ & \big| \widetilde{\alpha}_i \big| \leq \eta, \quad i \in \{1,\ldots,n\},
\end{cases}
\label{eq:KSVM}
\end{equation}

where  $\widetilde{\mathbb{G}} = U\,S\,\Lambda\,U^{\tt t} $ with  $U$ and  $\Lambda$ the eigenvector and eigenvalue matrices of $\mathbb{G} = U\,\Lambda\,U^{\tt t}$, and $S = \text{sign}(\Lambda) $. Since problem (\ref{eq:KSVM}) is the same as (\ref{eq:SVM}), it is immediate to see that it can be solved using a standard SVM solver on $\widetilde{\mathbb{G}}$.

\end{document}